\documentclass[lettersize,journal]{IEEEtran}
\usepackage{amsmath,amsfonts}
\usepackage{algorithmic}
\usepackage{algorithm}
\usepackage{array}
\usepackage[caption=false,font=normalsize,labelfont=sf,textfont=sf]{subfig}
\usepackage{textcomp}
\usepackage{url}
\usepackage{verbatim}
\usepackage{graphicx}
\usepackage{cite}
\usepackage{times}
\usepackage{latexsym}
\usepackage{longtable}
\usepackage{afterpage}
\usepackage{multibib}
\newcites{languageresource}{Language Resources}
\usepackage{graphicx}
\usepackage{tabularx}
\usepackage{soul}
\usepackage{lingmacros}
\usepackage[justification=centering]{caption}
\usepackage{times}
\usepackage{latexsym}
\usepackage{graphicx}
\usepackage{booktabs,multirow}
\usepackage{amssymb}
\usepackage{tabularx}
\usepackage{bm}
\usepackage{amsmath,mathtools}
\usepackage{amsfonts}
\usepackage{longtable}
\usepackage{wrapfig}
\usepackage{enumitem}
\usepackage{pifont}
\usepackage{longtable}
\usepackage{mwe}
\usepackage{makecell}
\usepackage{dblfloatfix}
\usepackage{xcolor}
\usepackage{siunitx}

\usepackage{varwidth}
\usepackage{array}

\usepackage{newfloat}
\usepackage{caption}
\usepackage{multicol}
\usepackage{float}

\usepackage[switch]{lineno}
\usepackage{lipsum}
\usepackage{epstopdf}
\usepackage[utf8]{inputenc}

\usepackage{hyperref}
\usepackage{xstring}

\usepackage{color}
\usepackage[T1]{fontenc}
\usepackage{graphicx}

\begin{document}

\title{From Words to Wisdom: Discourse Annotation and Baseline Models for Student Dialogue Understanding}


\author{Farjana Sultana Mim,
        Shuchin Aeron, 
        Eric Miller and
        Kristen Wendell%
\thanks{F. S. Mim is with the Department of Electrical and Computer Engineering, Tufts University, Medford, MA 02155, United States. (e-mail: farjana.mim59@gmail.com) (currently in the Department of Computer Science and Information Technology, Patuakhali Science and Technology University, Bangladesh)} 
\thanks{S. Aeron is with the Department of Electrical and Computer Engineering, Tufts University, Medford, MA 02155, United States. (e-mail: shuchin.aeron@tufts.edu) }
\thanks{E. Miller is with the Department of Electrical and Computer Engineering, Computer Science and Biomedical Engineering, Tufts University, Medford, MA 02155, United States. (e-mail: eric.miller@tufts.edu)}
\thanks{K. Wendell is with the Department of Mechanical Engineering and Education, Tufts University, Medford, MA 02155, United States. (e-mail:  kristen.wendell@tufts.edu)}
\thanks{}}




\maketitle

\begin{abstract}
Identifying discourse features in student conversations is quite important for educational researchers to recognize the curricular and pedagogical variables that cause students to engage in constructing knowledge rather than merely completing tasks.
The manual analysis of student conversations to identify these discourse features is time-consuming and labor-intensive, which limits the scale and scope of studies.
Leveraging natural language processing (NLP) techniques can facilitate the automatic detection of these discourse features,
offering educational researchers scalable and data-driven insights.
However, existing studies in NLP that focus on discourse in dialogue rarely address educational data.
In this work, we address this gap by introducing an annotated educational dialogue dataset of student conversations featuring knowledge construction and task production discourse. 
We also establish baseline models for automatically predicting these discourse properties for each turn of talk within conversations, using pre-trained large language models GPT-3.5 and Llama-3.1.
Experimental results indicate that these state-of-the-art models perform suboptimally on this task, indicating the potential for future research.
\end{abstract}


\section{Introduction}
\IEEEPARstart{R}{esearch} in classroom settings has shown that student learning outcomes are higher when students frame a classwork or homework activity as an opportunity for constructing knowledge rather than as a task to be produced for the instructor \cite{gouvea2019epistemological, koretsky2014productively}. 
In other words, two important features of student conversations are: \emph{knowledge construction (KC)} discourse, which refers to the student talks focused on developing conceptual understanding, and \emph{task production (TP)} discourse, where student talks are focused on completing an instructional task as expediently as possible \cite{swenson2018developing}.

Prior research in learning sciences has also demonstrated that when students frame their purpose within an instructional activity as constructing knowledge rather than just completing a task, they are more likely to develop expertise and be able to later transfer their expertise to new situations \cite{schwartz2016abcs}.
These results have been found across several disciplines, including physics, chemistry, biology, and engineering education \cite{gouvea2019epistemological, koretsky2014productively, scherr2009student, swenson2018developing}.
However, the relationship between knowledge construction discourse and learning outcomes has yet to be translated into actionable principles for pedagogy and curriculum design. 
The major difficulty lies in pinpointing which particular aspects of the learning environment and instructional activity cue students into knowledge constructing discourse.

To address this gap, we aim to develop efficient methods for distinguishing students’ knowledge construction discourse from their task production discourse so that researchers can more broadly investigate the conditions or contexts under which students tend to adopt a knowledge construction framing. 
Such findings would enable educators to design learning experiences and environments so that they cue students toward constructing knowledge.

\begin{figure}[t]
    \small
    \begin{tabular}{p{0.94\linewidth}}
        \toprule
        \textbf{Homework Topic} \\
        \midrule
        Design an experiment complete with instrumentation to determine the specific heats of a gas using a resistance heater. Discuss how the experiment will be conducted, what measurements need to be taken, and how the specific heats will be determined. What are the sources of error in your system? How can you minimize the experimental error? \\
        \midrule
        \midrule
        \textbf{Task Production Discourse} \\
        \midrule
        \emph{Student X}: Although we just have to design the experiment. It’s not like we have to actually do it. \\
        \emph{Student T}: No. \\
        \emph{Student A}: Just design and justify this will work.\\
        \emph{Student X}: How can you minimize the experimental error. That’s one of the points there. \\
        \midrule
        \textbf{Knowledge Construction Discourse} \\
        \midrule
        \emph{Student X}: Ok. So one thought I had too was that actually um whatever material the container is made out of when it heats up, it’s going to expand - \\
        \emph{Student T}: Mhm. \\
        \emph{Student X}: - and that will change whatever the internal volume is. And I don’t know if it makes it bigger or smaller actually. It um might make it bigger but if there were - \\
        \bottomrule
    \end{tabular}
    \caption{Students' homework discussion's snippet of knowledge construction and task production discourse.}
    \label{fig:example}
\end{figure}

\IEEEpubidadjcol
Fig \ref{fig:example} shows examples of \emph{knowledge construction} and \emph{task production} discourse in an undergraduate engineering students' conversation.
In the task production discourse of the example, the students remind each other that their homework task is to design an experiment and describe how they would minimize experimental error. 
These lines are focused on setting up the steps to complete their homework. 
In the knowledge construction discourse from the same homework conversation, \emph{student X} shares an idea about how the process of heating a gas will affect the material containing it. Rather than simply completing a pre-determined step of the homework, student X tries to envision the phenomena that will occur in the experiment the students are designing. 
At this moment, X's turn of talk is oriented toward understanding rather than expediency.

Traditional manual analysis of student dialogues to identify these discourse features is time-intensive, which limits the scope of studies.
Leveraging natural language processing (NLP) techniques can facilitate the automatic detection of KC and TP discourse, providing educational researchers with valuable insights into how curricular and pedagogical variables influence students to engage in knowledge construction rather only task production.

Discourse in dialogue or conversations has been widely studied in NLP in different task settings such as dialogue act classification \cite{he2021speaker, raheja2019dialogue, li2018dual, kumar2018dialogue}, dialogue topic segmentation and categorization \cite{liu2023joint, xing2021improving, somasundaran2020two, kim2015towards}, dialogue state tracking \cite{feng2023towards, xu2023dialogue, ma2023parameter, guo2022beyond, zhou2022dialogue, qixiang2022exploiting}, and identifying dialogue system behaviors \cite{finch2023leveraging, sabour2022cem}.
However, although various discourse frameworks are being applied to different types of conversational data, hardly any of them are educational data \cite{jensen2021deep, alic2022computationally}.
To address this gap, this study creates a novel educational dialogue dataset, annotated with knowledge construction (KC) and task production (TP) discourse\footnote{The human subjects protocol under which the data were generated does not allow for its public sharing. Readers interested in the data set may contact the authors for further information.}. 
We also formulate the NLP task of KCTP (Knowledge Construction and Task Production) prediction, aiming to automatically identify these discourse types within educational dialogues.

Lately, the NLP field has been revolutionized by pre-trained large language models (LLMs) such as GPT-3 \cite{brown2020language}, Llama \cite{touvron2023llama}, Gemini \cite{reid2024gemini}, Deepseek \cite{guo2025deepseek}.
These models have demonstrated significant performance gains and yielded interesting findings across various NLP tasks, including the study of discourse in dialogues or conversations \cite{feng2023towards, finch2023leveraging}. 
Recently, a new paradigm called \emph{``Pre-train, Prompt, and Predict''} \cite{liu2023pre} has gained popularity which leverages pre-trained LLMs through natural language prompts instead of fine-tuning them for specific tasks.
By using such \emph{``prompting''} method, one can probe task-specific knowledge from LLMs, which has shown remarkable performance in various tasks such as text classification and summarization \cite{gao2020making, li2021prefix}. Another paradigm called \emph{``instruction fine-tuning''} \cite{wei2021finetuned} which finetunes a model on a dataset via instructions, has significantly improved the performance of several tasks \cite{chung2024scaling}.
Therefore, we use GPT-3.5 with prompting techniques to establish a baseline for our Knowledge Construction vs. Task Production (KCTP) prediction task. However, as GPT-3.5 is not an open-source model, we also use the open-access LLaMA-3.1 (8B) model \cite{dubey2024llama} and fine-tune it for the same task. Experimental results indicate that prompting and fine-tuning GPT-3.5 and LLaMA-3.1 yield suboptimal performance on KCTP prediction, suggesting the need for further research into models and methods better suited to educational discourse analysis.
To summarize, the main contributions of this work are as follows:

\begin{itemize}
\item{We create a novel educational dialogue dataset annotated with \emph{Knowledge Construction (KC)} and \emph{Task Production (TP)} discourse, addressing a gap in discourse-annotated educational data.}
\item{We formulate the \emph{Knowledge Construction vs. Task Production (KCTP)} classification as a natural language processing (NLP) task to automatically identify \emph{KC} and \emph{TP} discourse in student dialogues.}
\item{We establish baseline models for the \emph{KCTP} prediction task using GPT-3.5 and LLaMA-3.1 prompting as well as LLaMA-3.1 instruction fine-tuning, revealing current limitations of LLMs in modeling educational discourse and highlighting directions for future research.}

\end{itemize}


\section{Related Work}
This study develops an educational dialogue dataset annotated with instances of \emph{knowledge construction (KC)} and \emph{task production (TP)} discourse. We also establish baseline models for the automatic prediction of KC and TP discourse, with the goal of enabling educational researchers to identify the curricular and pedagogical conditions that encourage students to engage in constructing knowledge rather than merely completing tasks.
In this section, we briefly review prior work in three relevant areas: (1) discourse in learning sciences, (2) discourse analysis in Dialogue using NLP, and (3) use of pre-trained language models for discourse modeling in dialogue.

\subsection{Learning Sciences Approach to Educational Discourse Analysis}
Discourse has been long studied in learning sciences to determine the nature of activity, understanding, and learning styles of students \cite{schwartz2016abcs, gouvea2019epistemological, koretsky2014productively, scherr2009student, swenson2018developing}.
 Gouvea et al. \cite{gouvea2019epistemological} presented a case study of a life-science major in a reformed physics course, showing how epistemological shifts in one discipline can transfer to another. Over a year, the student moved from rote learning to coherence-seeking reasoning in physics, integrating materials, peer discussion, and feedback. This reframing extended to biology, where the student began approaching the subject more conceptually. The study provides qualitative evidence that discourse-centered instructional strategies can foster cross-disciplinary epistemological development.
 
In another work, Scherr and Hammer \cite{scherr2009student} explored how students’ collaborative behaviors such as posture, gaze, gestures, and vocal dynamics serve as observable indicators of their epistemological framing during active-learning physics activities. They analyze video recordings from introductory physics tutorial sessions and identify distinct behavioral clusters corresponding to different ways students frame the task: for instance, working through substance-based sensemaking versus perceiving it as a procedural worksheet exercise. The authors demonstrate that when students frame the activity as sensemaking, their behaviors align with deeper conceptual reasoning and engagement in discussing the substance of ideas. Their findings highlight the dynamic interplay between observable behavior, framing, and the quality of students’ scientific reasoning in small-group learning contexts.

Koretsky et al. \cite{koretsky2014productively} examined how the design of engineering tasks and instructional framing influence student team dynamics, balancing action (“doing”) and reflection (“thinking”) during collaborative open-ended projects. Through detailed cases of small-group engineering design work, they show that when tasks are meaningful, realistic, and properly scaffolded, teams display more equitable participation, distributed modeling and communication, and deeper conceptual reasoning rather than surface-level task execution alone. In particular, the interplay between material engagement (e.g., prototyping and sketching) and explicit discourse about design decisions fosters collective sense-making and shared agency. The study highlights how thoughtfully structured activities and facilitative framing can empower teams to engage in both productive action and epistemic dialogue, offering important implications for discourse-centric analyses and NLP applications in educational dialogue modeling.

\subsection{Discourse Analysis in Dialogue using NLP}
Discourse in dialogue has been extensively studied in natural language processing (NLP) \cite{li2023task, tulpan2023deeper, mim-etal-2022-lpattack}. 
Raheja and Tetreault \cite{raheja2019dialogue} proposed a hierarchical recurrent neural network and coupled it with a context-aware self-attention mechanism to model different levels of utterance and dialogue act semantics, achieving state-of-the-art performance on the Switchboard Dialogue Act Corpus.
Liu et al. \cite{liu2023joint} introduced a joint model for dialogue segmentation and topic categorization, which was evaluated on a clinical spoken conversation dataset created by them.
In another work, Xu et al. \cite{xu2023dialogue} developed a Dialogue State Distillation Network (DSDN), which leverages relevant information of previous dialogue states and employs an inter-slot contrastive learning loss to effectively capture the slot co-update relations from dialogue context. Their proposed method achieved state-of-the-art performance on the dialogue state tracking task.
Sabour et al. \cite{sabour2022cem} introduced a novel approach for empathetic response generation in dialogue, which leverages commonsense
to draw more information about the user’s situation and uses
that to further enhance the empathy expression in generated responses. They showed that their approach outperforms the baseline models in both automatic and human evaluations.

\subsection{Use of Pre-trained Language Models for Discourse Modeling in Dialogue}

Recent advancements of pre-trained large language models (LLMs) has significantly advanced the field of discourse modeling \cite{li2024dialogue, cimino2024coherence, gu2021dialogbert, mim2021corruption}.
The importance of modeling speaker turns in dialogues was investigated by  He et al. \cite{he2021speaker}, where they incorporated turn changes in conversations among speakers for the dialogue act classification task. They introduced speaker turn embeddings and added them to utterance embeddings produced by the pretrained language
model RoBERTa \cite{liu2019roberta}, which showed better performance for the dialogue act classification task.
Xing and Carenini \cite{xing2021improving} utilized a neural utterance-pair coherence scoring model based on fine-tuning NSP BERT \cite{devlin2019bert} and achieved state-of-the-art results on the Dialogue topic segmentation task across three public datasets.
Feng et al. \cite{feng2023towards} presented the first evaluation of ChatGPT on the dialogue state tracking task, highlighting its superior performance over prior methods. They also proposed an LLM-driven dialogue state tracking framework based on smaller, open-source foundation models and showed that it achieves comparable performance to ChatGPT.
Finch et al. \cite{finch2023leveraging} investigated the ability of the state-of-the-art large language model (LLM),i.e., ChatGPT-3.5, to perform dialogue behavior detection for nine categories in real human-bot dialogues and showed that although ChatGPT performed promisingly, often outperforming specialized detection models, the result is still not up to human performance.

\begin{figure}[!t]
  \includegraphics[width=\columnwidth]{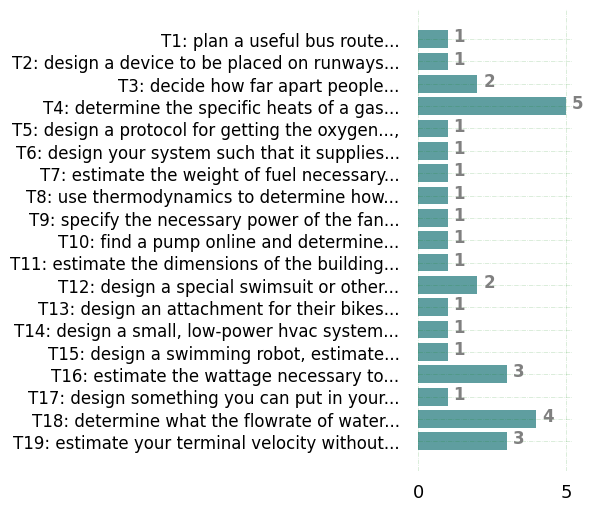}
  \caption{Topic distribution across the dataset}
  \label{fig:topic_distribution}
\end{figure}

Few researches have been conducted that focus on discourse modeling on educational dialogue data. 
Jensen et al. \cite{jensen2021deep} proposed a methodology for providing teachers with objective, automated feedback on the quality of their classroom discourse by comparing traditional open-vocabulary approaches using n‑grams and Random Forest classifiers with a modern deep transfer learning method leveraging BERT. By modeling seven key features of teacher talk (such as questioning and elaborated evaluation) on 127 recordings of classroom talk, the authors demonstrated that while transfer learning with BERT offers a promising path for enhancing automated discourse analytics in education, its effectiveness hinges on the availability of sufficient annotated data to fine-tune the model effectively.
Alic et al. \cite{alic2022computationally} automatically distinguished between two pedagogically significant types of teacher questions: funneling questions, which guide students toward specific answers, and focusing questions, which encourage students to reflect on their reasoning. The authors create a labeled dataset of over 2,000 teacher questions annotated by experts and develop both supervised (fine-tuned RoBERTa) and unsupervised models to classify question types. Their supervised RoBERTa model showed strong alignment with expert judgments and correlates with key educational outcomes, such as instructional quality and student learning gains.

\section{Dataset Curation}

\subsection{Data Collection}
We recorded homework discussions among undergraduate mechanical engineering students, focusing on topics from their thermal fluid systems course.
Between 2 and 5 students participated in each conversation.
Then, we transcribed the conversations ensuring that all data were de-identified. 
As part of the consent process, students were asked if their de-identified transcripts could be used in future research. Only transcripts from students who consented were included in the dataset.

\subsection{Dataset Statistics}
The dataset consists of 32 small-group conversations covering 19 homework topics, each topic corresponding to a distinct \emph{task description} that students were required to complete collaboratively through discussion.
Fig \ref{fig:topic_distribution} shows the topic distribution across the dataset.
The utterances in the conversations are segmented based on the fact that one “turn,” or utterance, consists of everything a single person utters until another person speaks (either because the first person has finished or because they interrupt the first person).
The average token per conversation is 6404, and the average turns of talk is 321.
Please see the appendix for the details of each topic.

\begin{figure}[!t]
  \centering
  \includegraphics[width=\columnwidth, height=6.5cm]{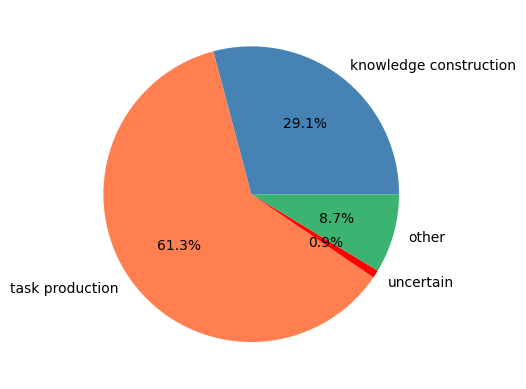}
  \caption{Distribution of categories across the dataset.}
  \label{fig:label-distribution}
\end{figure}

\subsection{Annotation Study}

\subsubsection{Setup}
Two expert annotators, including one co-author of this paper, participated in the annotation study. 
We developed a comprehensive annotation guideline and instructed the annotators to label each conversational turn as \emph{knowledge construction}, \emph{task production}, \emph{uncertain}, or \emph{other}. 
We created the label \emph{uncertain} for the turns of talk where there is insufficient evidence to determine whether the speaker is continuing the current framing of either knowledge construction or task production.
If a single utterance includes indicators for both KC and TP classification, and the annotator cannot determine which category is the predominant framing for the student during the utterance, the utterance should be classified as \emph{uncertain}.
The label \emph{other} refers to the turns of talk where students discuss a topic other than the assigned problem, such as the purpose of participating in the research study, or other academic classes or social events.


We trained the annotators in a pilot annotation phase where they were asked to annotate 5 conversations. After the pilot annotation, we discussed the disagreements and, if needed, adjusted the annotation guidelines.
In our main annotation study, 6 conversations were annotated by two annotators and 21 conversations were annotated by a single annotator.
For inter-annotator agreement (IAA) and the analysis of annotations, we report the results of dual annotations.
Fig \ref{fig:label-distribution} illustrates the distribution of annotated labels across the dataset of 32 conversations.
As anticipated, we see that the dataset is imbalanced, with the majority of annotated labels falling into the \emph{knowledge construction} and \emph{task production} categories.

\begin{figure}[!t]
  \includegraphics[width=\columnwidth]{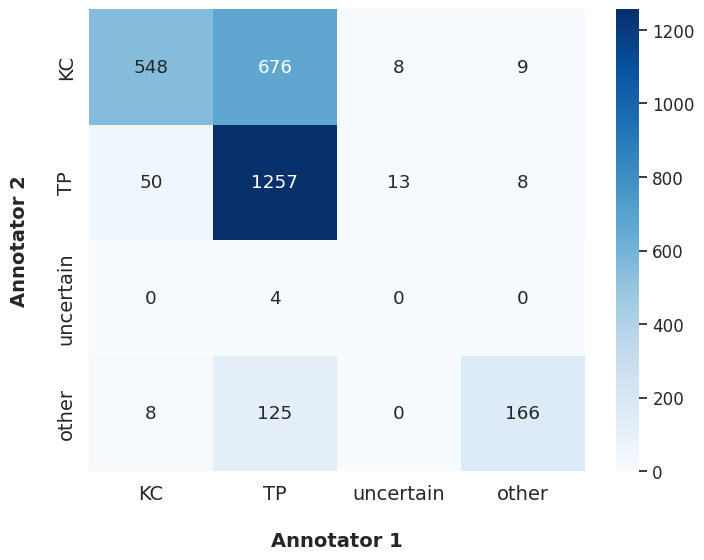}
  \caption{Confusion matrix of dual annotations}
  \label{fig:annotator-disagreement}
\end{figure}

\subsubsection{Inter-annotator agreement (IAA)}
We computed IAA using Cohen’s ($\kappa$) \cite{cohen1960coefficient} for the dual annotations across four annotated discourse labels (i.e., knowledge construction, task production, uncertain and other) .
We obtained Cohen’s ($\kappa$) of 0.45 which indicates a moderate agreement \cite{artstein2008inter, spooren2010coding}.

Discerning undergraduate students’ aims and purposes based on their spoken word is notoriously difficult for learning sciences researchers \cite{scherr2009student}.
The difficulty in quickly determining whether students are in task production or knowledge construction mode (or when those modes are co-occurring) is one reason researchers are interested in exploring algorithm-assisted annotation. This also means that it is not surprising that the agreement between the two annotators was only moderate.


\subsubsection{Analysis of Annotations}
The confusion matrix of the dual annotations of 6 conversations is shown in Figure \ref{fig:annotator-disagreement}.
We see that both annotators mostly agree with each other during the labeling of \emph{task production (TP)} discourse and the most disagreement happens when one annotator thinks a turn of talk is \emph{knowledge construction (KC)} while other thinks that it's \emph{task production}.

Figure 4 reveals that Annotator 1 leaned toward classifying discourse as TP, while Annotator 2 leaned toward classifying discourse as KC. 
Of the utterances on which there was TP vs KC disagreement between the two human annotators, Annotator 1 chose KC for only 7\% (50/726) of the disagreements while Annotator 2 chose KC in 93\% (676/726) of the cases.
Besides, where there was TP vs “other” disagreement, 94\% (125/133) times Annotator 1 chose TP over other, while just 0.06\% (8/133) times Annotator 2 chose TP over “other.”


We also found that the disagreement mostly happens under two conditions: (i) when students discuss the details of their problem-solving steps, and (ii) when students ask each other questions. 
For example, Consider the conversation snippet below (from Topic 4, ``determine the specific heats'').
\begin{itemize} [label={}]
    \item \emph{T: Right. Also how long are we doing it for. It's for like - }

    \emph{X: Yeah. Do it for ten hours. Do we need another you know
    microsecond.}
    
    \emph{T: Yeah. Um ok. So then graph um V I versus time and take the area under the curve. Um. Ok. That area under the curve is just gonna be equal to Q, right?}
\end{itemize}

The students here are discussing the details of an experimental design. Their homework task is to specify the design set-up. They consider the duration of the experiment, the plot they will produce from the data, and the physical quantity represented on that plot. 
Annotators 1 and 2 disagree on whether this portion of the discussion is aimed toward deeper understanding or toward making progress on the assignment. 
On one hand, discussion of experiment timescales and of the meaning of a graph might help students build knowledge about the physical quantity to be measured in the experiment. On the other hand, the students’ statements about the length of the experiment and of the graphs it will generate could simply comprise another step forward in specifying an experimental design, which is completing the homework task.

The correct label in the case is knowledge construction. When student X discusses the experiment’s timescale and student T discusses the meaning of the area under the curve, they are talking about concepts that they contributed anew to this homework discussion; these were not concepts mentioned in the homework problem statement, notes, or textbook for this course. Therefore, the students were calling up other intellectual resources to construct new ideas for this homework activity.

\section{Experiments}

\subsection{Task setting}
We consider the prediction task of KCTP discourse in a conversation as a label-generation task for each turn of talk in the conversation, where the model is instructed to generate one label out of the four annotated labels i.e., \emph{knowledge construction (KC), task production (TP), uncertain, other} for each turn of talk. 

To create a strong baseline, we assume that in cases where such KCTP discourse-specific resources are unavailable, pre-trained large language models (LLMs) could be the most effective means of generating KCTP discourse labels for each turn in the conversation. 
We evaluate our task in three settings: (i) \textbf{Zero-shot prompting setting}: Zero-shot prompting is a technique used with large language models in which a task is defined using only natural language instructions, without providing any examples of how the task should be performed.
This method relies on the model’s pre-trained knowledge and ability to generalize in order to accurately interpret and carry out the given instruction.
(ii) \textbf{Few-shot prompting setting}: Few-shot prompting is a technique where a language model is given a few input-output examples along with a natural language instruction to guide its response to new, similar inputs.
In contrast to zero-shot prompting, which relies only on instructions, few-shot prompting uses these examples to establish a pattern or context that the model can mimic.
This method exploits the in-context learning ability of large language models, allowing them to generalize from a small number of examples without the need for task-specific fine-tuning \cite{brown2020language}.
(iii) \textbf{Fine-tuning setting}: Fine-tuning refers to the process of taking a pre-trained large language model (which is generally trained on a large, general-purpose corpus) and further training it on a smaller, task-specific dataset to improve its performance on a particular task. This transfer learning strategy \cite{devlin2019bert} allows models to leverage the rich representations learned during pre-training and adapt them to specialized tasks.

\begin{table*}[!t]
  \centering
  \small
  \begin{tabular}{c|ccccccccc}
    \toprule
    \multirow{3}{*}{\textbf{Prompts}} &
    \multicolumn{4}{c}{\textbf{Zero-Shot}} & \multicolumn{4}{c}{\textbf{Few-Shot}} & \multicolumn{1}{c}{\textbf{Fine-Tuned}}\\
    \cmidrule{2-10}
    & \multicolumn{2}{c}{Curated prompt} 
    & \multicolumn{2}{c}{Optimized prompt} & \multicolumn{2}{c}{Curated prompt} 
    & \multicolumn{2}{c}{Optimized prompt} & \multicolumn{1}{c}{Optimized prompt}\\
    \cmidrule{2-10}
    & GPT-3.5& Llama-3.1 & GPT-3.5 & Llama-3.1  & GPT-3.5 & Llama-3.1 & GPT-3.5 & Llama-3.1 & Llama-3.1 \\
    \hline
    Prompt 1   & 0.34 & 0.43 & 0.29 & 0.49 & 0.35 & 0.48 & 0.26 & 0.50 & 0.54  \\
    Prompt 2   & 0.28 & 0.38 & 0.47 & 0.40 & 0.35 & 0.45 & 0.37 & 0.48 &  0.51                   \\
    Prompt 3   & 0.49 & 0.47 & 0.55 & 0.44 & 0.39 & 0.51& 0.40 & 0.47 & 0.45                     \\
    Prompt 4   & 0.32 & 0.52 & 0.46 & 0.50 & 0.38 & 0.54& 0.44 & 0.57 & 0.49                  \\
    Prompt 5   & 0.27 & 0.39 & 0.33 & 0.46 & 0.27 & 0.44 & 0.27 & 0.49 & 0.55 \\
    \bottomrule
  \end{tabular}
  \caption{
    Performance of GPT-3.5 and Llama-3.1 in the label prediction task under \\ zero-shot, few-shot, and fine-tuned settings using different prompt types.}
  \label{table:model-performance}
  
\end{table*}

\subsection{Models}
We employ state-of-the-art LLMs namely GPT-3.5-turbo \cite{brown2020language} and Llama-3.1-8B-Instruct \cite{dubey2024llama} models for the KCTP discourse prediction task while we use GPT-4-1106-preview \cite{achiam2023gpt} for our prompt engineering \cite{liu2023pre}. 
A GPT (Generative Pre-trained Transformer) model \cite{radford2018improving} is an  auto-regressive large language model developed by OpenAI that uses transformer \cite{vaswani2017attention} architecture to generate and understand human-like text.
GPT models use a transformer decoder architecture, which is trained to predict the next word in a sequence, followed by fine-tuning on labeled datasets for specific applications.
GPT-3.5 Turbo is optimized for speed and cost-efficiency, making it ideal for high-volume tasks. In contrast, GPT-4 offers superior reasoning, accuracy, and contextual understanding for more complex applications while costing more as well.
Like GPT, the LLaMA (Large Language Model Meta AI) series \cite{touvron2023llama} developed by Meta is also an auto-regressive language model based on the transformer architecture. Its key advantage lies in being open-source, enabling cost-free use while still delivering competitive performance.

\subsection{Prompt Design}
We create 5 prompts for the KCTP prediction task and use the GPT-4-1106-preview model to optimize our created prompts. We report results for both types of prompts, i.e., our curated prompts and GPT-4 optimized prompts.
We also use the optimized prompts for instruction fine-tuning of Llama-3.1 8B model.
Among the 5 prompts, prompts 1 and 2 consist of the previous dialogue context along with the current turn of talk.
Prompts 3 and 4 include the task description and the definitions of the labels respectively in addition to the previous dialogue context and current turn of talk.
Prompt 5 consists of both the previous and afterward dialogue context and the current turn of talk.
Please see the details of these prompts in Appendix.

\subsection{Setup}
We conduct experiments in zero-shot and eight-shot (two examples for each of the four labels) prompt settings where the number of shots reflects the number of examples provided in the prompt. Few-shot examples were sampled from two conversations and topics not included in the dataset.

We use OpenAI's API for GPT models and set the temperature of the model at 0.0 and maximum tokens at 10.
To use Llama-3.1 (8B) for our task, we use \emph{Unsloth}, an open-source AI library that enables us to train an LLM faster and efficiently with less GPU memory by applying techniques like quantization \cite{jacob2018quantization} and low-rank adaptation (LoRA) \cite{hu2022lora}.
In our zero-shot and few-shot experiments, we set the Llama-3.1 model with a temperature of 1.5 and a maximum of 64 new tokens. Fine-tuning is performed for 5 epochs using a learning rate of 1e-4 with the AdamW 8-bit optimizer. We use a batch size of 8, gradient accumulation of 16, and a weight decay of 0.01. All experiments are conducted with a fixed random seed of 3407.

\subsection{Evaluation Procedure}
We use the weighted F1 score to evaluate model performance. During fine-tuning Llama-3.1, we employ five-fold cross-validation to obtain results across the entire dataset and enable comparison with zero-shot and few-shot prompting.

\begin{table*}[!t]
  \centering
  \begin{tabular}{c||c|c|c}
    \toprule
    \textbf{Prompts} & \textbf{Same topic} & \textbf{Single different topic} & \textbf{Three different topics}\\
    \hline
    Prompt 1   & 0.64 & 0.61 & 0.65      \\
    Prompt 2   & 0.58 & 0.63 & 0.61                       \\
    Prompt 3   & 0.60 & 0.63 &  0.61                         \\
    Prompt 4   & 0.53 & 0.58 & 0.62                     \\
    Prompt 5   & 0.64 & 0.62 &  0.61    \\
    \bottomrule
  \end{tabular}
  \caption{
    Performance of finetuned Llama-3.1 when trained on the same topic, \\ a single different topic, and three different topics.
  }
  \label{table:model-performance-on-different topics}
\end{table*}

\section{Results and Discussions}
Table \ref{table:model-performance} presents the performance of GPT-3.5 and LLaMA-3.1 models across three experimental settings: Zero-Shot, Few-Shot, and Fine-Tuned, using both curated and GPT-4 optimized prompts on the KCTP discourse label prediction task.
Five prompts (Prompt 1–5) developed by us are evaluated, and the scores represent the average F1 score.


\subsection{Zero-shot and Few-shot effectiveness}
The results indicate that overall, Llama-3.1 performs better than GPT-3.5 for both curated and optimized prompts across zero-shot and few-shot settings. 
However, the overall performance remains suboptimal, with the highest F1-score reaching only 0.57.

In the zero-shot setting, the best result is obtained by GPT-3.5 with optimized prompt 3, which includes a topic description in addition to the previous dialogue contexts. The result suggests that explicitly providing the communicative goal of the student conversation helps the model infer appropriate labels without prior examples.
However, the GPT-3.5 performance drops in the few-shot setting. We assume that one of the reasons the few-shot prompting did not perform better here could be attributed to the fact that the examples we used didn't generalize well with the dataset, or the model had too much information to process.
Moreover, performance degradation can sometimes happen in some LLMs for adding too domain-specific examples \cite{tang2025few}.
In contrast, LLaMA-3.1 attains its highest score in the few-shot setting with optimized prompt 4, which incorporates label definitions alongside the preceding dialogue context. It means that when we include examples, LLaMA-3.1 can better generalize than GPT-3.5 on this task by leveraging explicit label information.
Table \ref{table:model-performance} also shows that optimizing our curated prompt with GPT-4 improves the overall performance. 

\begin{figure}[!b]
  \includegraphics[width=\columnwidth]{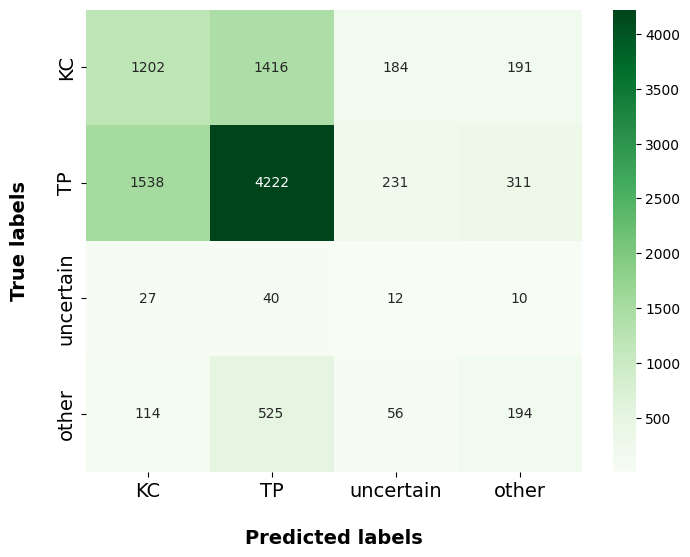}
  \caption{Confusion matrix of model prediction \\ vs. true annotated labels}
  \label{fig:prediction-matrix}
\end{figure}


\subsection{Fine-tuning effectiveness}

The fine-tuning results presented in Table \ref{table:model-performance} show that fine-tuning Llama-3.1 does not exceed its zero-shot and few-shot performance. 
We hypothesize that since our data is very domain-specific, it may have reduced the model’s generalization ability by overfitting to narrow linguistic or conceptual patterns. Recent studies also showed that, in certain cases, fine-tuning often yields limited or even negative performance gains for large language models (LLMs) \cite{barnett2024fine}.

Furthermore, the results suggest that incorporating dialogue context alone (as in Prompts 1 and 3), without additional task-specific contextual signals, leads to relatively better outcomes during fine-tuning. This observation implies that minimal yet coherent contextual grounding may help the model retain its pretrained reasoning and discourse capabilities.


\subsection{Analysis}
Figure \ref{fig:prediction-matrix} presents the confusion matrix of the true annotated labels versus the model-predicted labels for the best result (i.e., Llama 3.1 performance in the few-shot settings with optimized prompt 4). 
It shows that the model is good at predicting the \emph{task production} labels. However, it struggles with predicting the \emph{knowledge construction} label and misclassifies about half of these instances as \emph{task production}, indicating notable room for improvement. 
Moreover, the model also kind of struggles with predicting the \emph{``other''} label, only correctly predicting it 22\% of the time.
These findings suggest that while the model captures dominant discourse functions well, it struggles with more nuanced or less frequent categories, highlighting opportunities for improving label representation and contextual modeling.


Since our dataset consists a diverse set of 19 topics (see Appendix for the details of each topic), we investigated how topic similarity between training and testing data influences model performance. For this experiment, we kept the test data same, and the training data size fixed, but varied the topical composition of the training data across three configurations: (i) training and test data drawn from the same topic, (ii) training data from a different single topic than the test set, and (iii) training data from three different topics distinct from the test set.

The results, summarized in Table \ref{table:model-performance-on-different topics}, reveal that, the model performs better when trained on data from different topics, except for prompt 5.
This finding suggests that exposure to a broader range of linguistic and conceptual patterns enhances the model’s generalization ability, whereas training on a single, homogeneous topic may lead to overfitting for most prompts. 
Notably, the model exhibits slightly better generalization within the same-topic setting for Prompt 5, which incorporates succeeding dialogue context.
This indicates that the model might learn topic-bounded discourse dynamics i.e., how participants introduce, elaborate on, and shift ideas within a coherent topical space in this prompt setting whereas such recurring patterns are less transferable across topics.





\section{Limitations}
In this study, we only consider two types of discourse features in educational conversations, namely \emph{knowledge construction (KC)} and \emph{task production (TP)}. 
Besides, the dataset is limited to topics from a thermal fluid systems course in mechanical engineering
Moreover, we use a limited set of prompt templates because of the resource and time constraints.

\section{Conclusions and Future Work}
This work presents a novel educational conversational dataset, annotated with \emph{knowledge construction (KC)} and
\emph{task production (TP)} discourse. 
Such discourse properties are crucial for framing student learning activities to develop more effective pedagogical settings that emphasize knowledge construction over mere task completion.
In this work, we establish baselines for the KCTP discourse prediction task using state-of-the-art language models with prompting techniques and fine-tuning.
Our results demonstrate that state-of-the-art LLMs struggle with this task, both under prompting-based and fine-tuning settings.

For future work, we plan to create reasoning chains that will help the model better understand the definitions of the labels in the prompts. We also intend to annotate low-level discourse structure for these student dialogues so that looking at the lower levels might help to see how the higher-order concepts emerge from a particular interaction of dialogue moves.
Furthermore, we aim to expand the dataset by including a broader range of undergraduate subjects, thereby capturing more diverse discourse patterns across academic domains. This increased topical diversity will support more robust fine-tuning and facilitate the development of models capable of domain-general discourse understanding.



\clearpage
\onecolumn
\appendix
\begin{longtable}{| p{.12\textwidth} | p{.82\textwidth}|}
\toprule
\centering
\small
\textbf{Topic} & \textbf{Description}  \\
\midrule
Topic 1 (T1) &  about a decade ago, stanford university successfully tried using waste vegetable oil from the dining halls as fuel for campus shuttles (https://news.stanford.edu/news/2006/january25/biodiesel-012506.html). what if [institution] tried to do this? plan a useful bus route around [institution] and specify the volume of fuel needed for the bus to travel this route without having to refuel. you may assume the energy density of vegetable oil is 42.20 mj/kg or 30.53 mj/l. \\
    \hline 
    Topic 2 (T2) & contrails are giant vortices left by airplanes on the runway and in the sky. if other planes pass through these, it can cause problems because it is like going through a mini tornado, and the planes are not equipped to handle such a pressure gradient. boeing has hired you to design a device to be placed on runways to help get rid of contrails there. this could be done by moving the contrails out of the way or by stopping them altogether. justify your design using fluid mechanics.\\ \hline 
    Topic 3 (T3) & covid-19 has drastically changed how people live their daily lives. guidelines have been created for how far apart people should stay when talking normally to each other. however, if people are doing something like singing, which takes more effort results in air (and droplets potentially carrying the coronavirus) being expelled from the lungs more forcefully, the guidelines for simply talking may not be adequate. if six feet apart is the recommendation for talking, use fluid mechanics argument to decide how far apart people who are singing should stand in order to be far enough away from any particles that may be expelled into the air by their singing. \\ \hline 
    
    Topic 4 (T4) & design an experiment complete with instrumentation to determine the specific heats of a gas using a resistance heater. discuss how the experiment will be conducted, what measurements need to be taken, and how the specific heats will be determined. what are the sources of error in your system? how can you minimize the experimental error? \\ \hline 
    
    Topic 5 (T5) & gas turbine engines used in airplanes consist of a fan followed by a compressor, diffuser, combustor, turbine, and sometimes an afterburner. you are designing the engine for a high- altitude airplane. normally, commercial planes operate best around 35,000 ft above sea level, but your plane should operate optimally at around 100,000 ft. because of the high altitude, there will be a lower concentration of oxygen than normal, and the air entering the engine will be colder. design a protocol for getting the oxygen up to the appropriate temperature and pressure needed for combustion. keep in mind your solution has to be relatively light. \\ \hline 
    
    Topic 6 (T6) & geothermal heat pumps harness renewable geothermal energy by using thermal reservoirs of water deep within the earth for heating.  such reservoirs have temperatures up to around 370 degrees celsius.  geothermal heat pumps use this energy by transporting room-temperature or cold liquid deep into the ground via pipes, exposing it to the hot reservoir, and carrying it back up to the surface.  imagine one of these reservoirs is discovered beneath the building where you live and design a heat exchanger system that uses the reservoir to heat your building.  sketch your system and specify the diameter, length, and material of the pipe, the flow rate, and the working fluid.  design your system such that it supplies a significant portion of the energy required for your building to operate normally. \\ \hline 
    
    Topic 7 (T7) & hybrid rockets use a combination of solid and liquid or gaseous propellants. in hybrid rockets, a stable oxidizer is used with a solid fuel. in order to be used, the fuel needs to be vaporized. the primary difficulty with hybrids is with mixing the propellants during the combustion process. in a hybrid rocket, the mixing happens at the melting or evaporating surface. the mixing is not well-controlled and generally, a lot of propellant is left unburned, limiting the motor’s efficiency. on the other hand, liquid propellants are generally mixed with oxidizer by an injector at the top of the combustion chamber which directs many small streams of fuel and oxidizer into one another. based on reasonable efficiencies of both liquid fuel and hybrid fuel processes, estimate the weight of fuel necessary to get a specific rocket of your choice to low earth orbit if the fuel is liquid vs. hybrid. \\ \hline 
    
    Topic 8 (T8) & most ski resorts in the u.s. use snow guns to make additional snow to supplement natural snow. these machines use water and compressed air. the air forces the water to form tiny droplets, which are then expelled from the nozzle and form ice crystals, which then fall to the ground as snow. compressed air cools as it expands, which assists with converting the water droplets into snow. choose your favorite ski resort and the desired depth of snow for the best skiing, and use thermodynamics to determine how long it will take to cover the ski trails in that amount of snow. you may assume that one snow gun uses about 100 gallons of water per minute and that the compressor can produce 50 cfm (cubic feet per minute) of air. \\ \hline 
    
    Topic 9 (T9) & race cars need to be as aerodynamic as possible. in many cases, to test the aerodynamics of a car, a wind-tunnel is used. you have been hired by chevrolet to analyze the air flow around their race cars. the wind tunnel you will be using to do this is an open circuit wind tunnel, where air is drawn from the laboratory environment, rather than being recirculated in the wind tunnel itself. such wind-tunnels consist of a nozzle to accelerate the air, the test section in which the car sits, and a diffuser which decelerates the air. based on reasonable values for air speed around the vehicle being tested, design a wind-tunnel for testing a race car. include all necessary specifications of the different parts of the wind tunnel, such as dimensions and air speeds.   also specify the necessary power of the fan and estimate the head loss due to the vehicle. use fluid mechanics to justify your response. \\ \hline 
    
    Topic 10 (T10) & since they know you are a mechanical engineer, your neighbors have asked you to help them design a waterfall for their garden similar to the one in the image below. you need to devise a way to get water from the pool at the bottom up to the top of the waterfall, and there needs to be enough water so that the waterfall actually looks good. design a system to do this. include a diagram of how the pump system will work, and include any important specifications such as flow rates and dimensions. then find a pump online and determine approximately how much power the waterfall pump will use per day. you may make as many assumptions as needed, just specify what assumptions you are making and why. \\ \hline 
    
    Topic 11 (T11) & the building of farfar’s danish ice cream shop in duxbury, ma is somewhat old and thus does not seem to have a great cooling system. as a result, sometimes the ice cream gets a bit melty even when it’s still in the freezer. the temperature in the ice cream shop is to be maintained at 55°f. estimate the dimensions of the building, use thermodynamics principles to determine the maximum heat loss the shop can have, and suggest a method for minimizing this heat loss. \\ \hline 
    
    Topic 12 (T12) & the butterfly swimming stroke is considered by many to be one of the most difficult strokes. it is also one of the fastest. when used over longer distances, the butterfly stroke is slightly slower than freestyle, partly due to the greater physical exertion required by the butterfly. however, butterfly has the fastest peak speed. explain why you think this stroke has the fastest peak speed. then, design a special swimsuit or other (non-motorized) device for a swimmer to further increase the speed of the butterfly stroke so that it will always be faster than freestyle no matter the distance over which the stroke is used. include a diagram of your design, and use fluid mechanics principles to prove that it will work. \\ \hline 
    
    Topic 13 (T13) & trek bikes has contracted you to design an attachment for their bikes to help make the bike and rider more streamlined. this attachment should effectively reduce the bike and rider’s air resistance without impeding the cyclist’s ability to ride their bike as usual. also specify what material this should be made of, and include a diagram of your design. justify your design using fluid mechanics. \\ \hline 
    
    Topic 14 (T14) & you are designing a tiny home that can be used for camping adventures. you want to be able to take your tiny home on camping trips in vermont and new hampshire during the fall to see the foliage, but you are worried that it might get a bit too cold for comfort, as that time of year, the temperatures at night can get down to \ang{30}F. design a small, low-power hvac system to keep the inside of your tiny home at a temperature no lower than \ang{45}F. specify what parts will be needed and how this system will be compatible with your tiny home. use fluid mechanics and heat transfer principles to justify that this system will indeed keep the temperature at \ang{45}F or higher. \\ \hline 
    
    Topic 15 (T15) & you are working at a robotics company to design a robot that can swim in water to collect data on sharks. this robot needs to be as hydrodynamic as possible so that it is efficient, and you need to be able to control how fast the robot will go so it can keep up with the sharks, as well as be able to make it turn while swimming. design a swimming robot, estimate its drag coefficient and the drag on the robot when it is moving at three different speeds (so you should have three different values for drag). then determine how much power will be needed to make the robot move forward at each of the three speeds. include a diagram of your robot in your response. \\ \hline 
    
    Topic 16 (T16) & you have been contracted by [institution] to design a system to get hot water to different parts of the science and engineering complex (sec). in particular, this system needs to work well during winter, when it is colder outside and most likely slightly colder than usual within the outer walls of the building and in the building in general. estimate the wattage necessary to keep the water at a reasonably hot temperature, and determine the flow rates and pressures necessary to get the hot water to various parts of the building. include a labeled sketch of your design, and be sure to use fluid mechanics to justify that your design will work. \\ \hline 
    
    Topic 17 (T17) & you have been doing a lot of baking recently and wish that you had a convection oven. convection ovens have one or more fans that help circulate the air in the oven, whereas in regular ovens, the only thing moving the air is natural convection. therefore, you want a convection oven so that you can bake everything faster and more evenly. however, you don’t want to spend the money on an entirely new oven since convection ovens are expensive, and you don’t want to have to get rid of the regular oven you already have. design something you can put in your regular oven that will make it function similarly to a convection oven. specify air flow rates and estimate the power needed for any components. also draw a diagram of your design and specify where any proposed components will go in the oven. use fluid mechanics to justify that your design will make your oven work similarly to a convection oven. \\ \hline 
    
    Topic 18 (T18) & you have been hired by firefighters to design a tripod to hold a large hose when fighting fires. the stream of water that comes out of the hose is 5 cm in diameter. determine what the flowrate of water out of the hose should be in order to work well for fighting a fire that is 9 meters away. then calculate how much reaction force will be needed at the base of the tripod to keep it from moving when the hose is being used. use fluid mechanics to support your response. \\ \hline 
    
    Topic 19 (T19) & you have recently gotten into skydiving. when you are skydiving, once you get close enough to the earth, you have to deploy a parachute. the skydiving part is exciting, but once you deploy the parachute, you have been getting bored since when you’re falling through the air, you eventually reach one constant speed (the terminal velocity). you want to design an attachment that enables you to increase and decrease your terminal velocity as you are falling. estimate your terminal velocity without this attachment, and then estimate the maximum and minimum terminal velocities when the attachment is being used. use lift and drag calculations to justify your answer.\\ \hline 
\caption{Details of the dataset topics, where each topic corresponds to a distinct \emph{task description} that students were required to complete collaboratively through discussion.} 
\label{tab:myfirstlongtable}
\end{longtable}

{
\small
\begin{longtable}{| p{.07\textwidth} | p{.45\textwidth} | p{.40\textwidth} |}
\toprule
\centering
\footnotesize
\textbf{Prompts} & \textbf{Author curated prompt template} & \textbf{GPT-4 optimized prompt template}\\
\midrule
  \textbf{Prompt 1} (Previous dialogue context)
  & You will be provided with a dialogue and its context.
  \vspace{.1cm}
  
  The context is the previous dialogue lines of the given dialogue and each line in context is separated by a newline character.
  \vspace{.1cm}
  
  Classify the given dialogue considering its context into one of the four categories: 
  knowledge construction, task production, uncertain, other. 
 \vspace{.1cm}
  
  Output only one of the categories and do not provide any explanation.
 \vspace{.1cm}
  
 \#\#\#\#
 \vspace{.1cm}
  
  Context:
\vspace{.1cm}
  
  Dialogue: 
  
  & Classify the provided dialogue into the correct category based on its context. Choose one of these categories: knowledge construction, task production, uncertain, or other. Only provide the category name as your response. 
    \vspace{.1cm}
    
    Context:
    \vspace{.1cm}
    
    Dialogue:\\
    
    \hline 
    
    \textbf{Prompt 2} (Previous dialogue context) & You will be provided with a current dialogue line and its previous dialogue lines.
    \vspace{.1cm}
    
    Each line in the previous dialogue lines is separated by a newline \textbackslash n character.
    \vspace{.1cm}

    Classify the current dialogue line considering its previous dialogue lines into one of the four categories: 
    knowledge construction, task production, uncertain, other
    \vspace{.1cm}

  Output only one of the categories and do not provide any explanation.
  \vspace{.1cm}

  \#\#\#\#
  \vspace{.1cm}

  Previous dialogue lines:
  \vspace{.1cm}

  Current dialogue line:
  \vspace{.1cm} 
  
  & Classify the current dialogue line into one of the following categories based on its context within the preceding dialogue lines: knowledge construction, task production, uncertain, or other. Provide the category without any explanation.
   \vspace{.1cm} 

  Previous dialogue context:
   \vspace{.1cm} 

  Current dialogue line:
   \vspace{.1cm}  
    
    \\ \hline
    \textbf{Prompt 3} (Previous dialogue context \& Topic description)
    & You will be provided with a dialogue, its context and a task description.
    \vspace{.1cm} 
    
  The context is the previous dialogue lines of the given dialogue and each line in context is separated by a newline character.
  \vspace{.1cm} 
  The dialogue and context are about completing the task details in the task description.
  \vspace{.1cm} 

  Classify the given dialogue considering its context and task description into one of the four categories: 
  knowledge construction, task production, uncertain, other.
  \vspace{.1cm} 

  Output only one of the categories and do not provide any explanation.
  \vspace{.1cm} 

  \#\#\#\#

  Task description:
  \vspace{.1cm} 

  Context:
  \vspace{.1cm} 

  Dialogue:
  \vspace{.1cm} 
  & Given a dialogue along with its preceding context and a specific task description, classify the provided dialogue into one of four categories (knowledge construction, task production, uncertain, other). Provide only the category without any further explanation. 
  \vspace{.1cm} 

  Task Description: 
  \vspace{.1cm} 

  Context: 
  \vspace{.1cm} 

  Dialogue:
  \vspace{.1cm} 

  \\ \hline
  \textbf{Prompt 4} (Previous dialogue context \& Label definitions) &
  You will be provided with a dialogue and its context.
  \vspace{.1cm}
  
  The context is the previous dialogue lines of the given dialogue, and each line in context is separated by a newline \textbackslash n character.
  \vspace{.1cm}
  Classify the given dialogue considering its context into one of the four categories: 
  knowledge construction, task production, uncertain, other.

  The definition of each of the categories is given below:
  \vspace{.1cm}
  
  knowledge construction: means the dialogue is focused on expressing understandings of concepts, phenomena, or technologies. Simply stating a definition from textbook or notes does not count as knowledge construction. 
  \vspace{.1cm}
  
  task production: means the dialogue is focused on completing the assigned task to satisfy the instructor, without verbalizing regard for understanding the bigger picture. For example, the dialogue is stating an equation, or asksing for a numerical answer, or calculating a number out loud, or discussing what to do next.
  \vspace{.1cm}
  
  uncertain: means there is insufficient evidence for classifying the dialogue either as a knowledge construction or task production. It is the default category for one word conversational fillers such as 'yeah', 'okay'.
  \vspace{.1cm}
  
  other: means the dialogue is about a topic other than the assigned task. 
  \vspace{.1cm}

  Output only one of the categories and do not provide any explanation.
  \vspace{.1cm}

  \#\#\#\#
  \vspace{.1cm}

  Context:
  \vspace{.1cm}

  Dialogue:
  \vspace{.1cm} & 
  
  Given the dialogue and its preceding context, classify the dialogue into one of the following four categories: knowledge construction, task production, uncertain, or other.
  \vspace{.1cm}

  - Knowledge Construction: The dialogue focuses on deep understanding of concepts or phenomena, going beyond mere definitions.
  \vspace{.1cm}
  
  - Task Production: The dialogue aims at completing a task or assignment, primarily focusing on procedural steps.
  \vspace{.1cm}
  
  - Uncertain: The dialogue does not provide enough information for classification into the above categories or includes filler words like 'yeah', 'okay'.
  \vspace{.1cm}
  
  - Other: The dialogue discusses topics unrelated to the assigned task.
  \vspace{.1cm}

  Provide only the category without any explanation.
  \vspace{.1cm}

  Context:
  \vspace{.1cm}

  Dialogue:
  \vspace{.1cm}
  \\ \hline

  \textbf{Prompt 5} (Previous and afterward dialogue context) & You will be provided with a dialogue, its before context and its after context \vspace{.1cm}
  \vspace{.1cm}
  
  The before context is the previous dialogue lines and after context is the succeeding dialogue lines of the given dialogue. Each line in before and after context is separated by a newline \textbackslash n character 
  \vspace{.1cm}

  Classify the given dialogue considering its before and after context into one of the four categories: 
  knowledge construction, task production, uncertain, other.
  \vspace{.1cm}

  Output only one of the categories and do not provide any explanation.
  \vspace{.1cm}

  \vspace{.1cm}
  \#\#\#\#

  \vspace{.1cm}
  Before Context:

  \vspace{.1cm}
  Dialogue:
  
  \vspace{.1cm}
  After Context:
  &
  Your task is to classify a specific dialogue based on the surrounding context into one of the following categories: knowledge construction, task production, uncertain, other. You will be given the dialogue, as well as the lines of conversation that precede it (Before Context) and follow it (After Context). Each dialogue line in the contexts is separated by a newline character. 

  \vspace{.1cm}
  Please provide only the category as your response without any explanation. 

  \vspace{.1cm}
  Before Context:

  \vspace{.1cm}
  Dialogue:

  \vspace{.1cm}
  After Context:

  \\ 
  \bottomrule

\caption{Details of the prompts used in modeling for knowledge construction and task production discourse prediction.} 
\label{tab:myfirstlongtable}
\end{longtable}}


 
%

\twocolumn
\bibliographystyle{IEEEtran}
\bibliography{ieee.bib}

@article{scherr2009student,
  title={Student behavior and epistemological framing: Examples from collaborative active-learning activities in physics},
  author={Scherr, Rachel E and Hammer, David},
  journal={Cognition and Instruction},
  volume={27},
  number={2},
  pages={147--174},
  year={2009},
  publisher={Taylor \& Francis}
}

@article{gouvea2019epistemological,
  title={Epistemological progress in physics and its impact on biology},
  author={Gouvea, Julia and Sawtelle, Vashti and Nair, Abhilash},
  journal={Physical Review Physics Education Research},
  volume={15},
  number={1},
  pages={010107},
  year={2019},
  publisher={APS}
}

@inproceedings{koretsky2014productively,
  title={Productively engaging student teams in engineering: The interplay between doing and thinking},
  author={Koretsky, Milo D and Gilbuena, Debra M and Nolen, Susan B and Tierney, Gavin and Volet, Simone E},
  booktitle={2014 IEEE Frontiers in Education Conference (FIE) Proceedings},
  pages={1--8},
  year={2014},
  organization={IEEE}
}

@phdthesis{swenson2018developing,
  title={Developing Knowledge in Engineering Science Courses: Sense-making and epistemologies in undergraduate mechanical engineering homework sessions},
  author={Swenson, Jessica Elizabeth Severson},
  year={2018},
  school={Tufts University}
}

@article{raheja2019dialogue,
  title={Dialogue act classification with context-aware self-attention},
  author={Raheja, Vipul and Tetreault, Joel},
  journal={arXiv preprint arXiv:1904.02594},
  year={2019}
}

@article{li2018dual,
  title={A dual-attention hierarchical recurrent neural network for dialogue act classification},
  author={Li, Ruizhe and Lin, Chenghua and Collinson, Matthew and Li, Xiao and Chen, Guanyi},
  journal={arXiv preprint arXiv:1810.09154},
  year={2018}
}

@inproceedings{kumar2018dialogue,
  title={Dialogue act sequence labeling using hierarchical encoder with crf},
  author={Kumar, Harshit and Agarwal, Arvind and Dasgupta, Riddhiman and Joshi, Sachindra},
  booktitle={Proceedings of the aaai conference on artificial intelligence},
  volume={32},
  number={1},
  year={2018}
}

@article{he2021speaker,
  title={Speaker turn modeling for dialogue act classification},
  author={He, Zihao and Tavabi, Leili and Lerman, Kristina and Soleymani, Mohammad},
  journal={arXiv preprint arXiv:2109.05056},
  year={2021}
}

@inproceedings{somasundaran2020two,
  title={Two-level transformer and auxiliary coherence modeling for improved text segmentation},
  author={Somasundaran, Swapna and others},
  booktitle={Proceedings of the AAAI Conference on Artificial Intelligence},
  volume={34},
  number={05},
  pages={7797--7804},
  year={2020}
}

@article{xing2021improving,
  title={Improving unsupervised dialogue topic segmentation with utterance-pair coherence scoring},
  author={Xing, Linzi and Carenini, Giuseppe},
  journal={arXiv preprint arXiv:2106.06719},
  year={2021}
}

@inproceedings{kim2015towards,
  title={Towards improving dialogue topic tracking performances with wikification of concept mentions},
  author={Kim, Seokhwan and Banchs, Rafael E and Li, Haizhou},
  booktitle={Proceedings of the 16th Annual Meeting of the Special Interest Group on Discourse and Dialogue},
  pages={124--128},
  year={2015}
}

@inproceedings{liu2023joint,
  title={Joint Dialogue Topic Segmentation and Categorization: A Case Study on Clinical Spoken Conversations},
  author={Liu, Zhengyuan and Salleh, Siti Umairah Md and Oh, Hong Choon and Krishnaswamy, Pavitra and Chen, Nancy},
  booktitle={Proceedings of the 2023 Conference on Empirical Methods in Natural Language Processing: Industry Track},
  pages={185--193},
  year={2023}
}

@article{feng2023towards,
  title={Towards LLM-driven dialogue state tracking},
  author={Feng, Yujie and Lu, Zexin and Liu, Bo and Zhan, Liming and Wu, Xiao-Ming},
  journal={arXiv preprint arXiv:2310.14970},
  year={2023}
}

@inproceedings{xu2023dialogue,
  title={Dialogue state distillation network with inter-slot contrastive learning for dialogue state tracking},
  author={Xu, Jing and Song, Dandan and Liu, Chong and Hui, Siu Cheung and Li, Fei and Ju, Qiang and He, Xiaonan and Xie, Jian},
  booktitle={Proceedings of the AAAI Conference on Artificial Intelligence},
  volume={37},
  number={11},
  pages={13834--13842},
  year={2023}
}

@article{ma2023parameter,
  title={Parameter-efficient low-resource dialogue state tracking by prompt tuning},
  author={Ma, Mingyu Derek and Kao, Jiun-Yu and Gao, Shuyang and Gupta, Arpit and Jin, Di and Chung, Tagyoung and Peng, Nanyun},
  journal={arXiv preprint arXiv:2301.10915},
  year={2023}
}

@article{guo2022beyond,
  title={Beyond the granularity: Multi-perspective dialogue collaborative selection for dialogue state tracking},
  author={Guo, Jinyu and Shuang, Kai and Li, Jijie and Wang, Zihan and Liu, Yixuan},
  journal={arXiv preprint arXiv:2205.10059},
  year={2022}
}

@inproceedings{zhou2022dialogue,
  title={Dialogue state tracking based on hierarchical slot attention and contrastive learning},
  author={Zhou, Yihao and Zhao, Guoshuai and Qian, Xueming},
  booktitle={Proceedings of the 31st ACM international conference on information \& knowledge management},
  pages={4737--4741},
  year={2022}
}

@inproceedings{qixiang2022exploiting,
  title={Exploiting domain-slot related keywords description for few-shot cross-domain dialogue state tracking},
  author={Qixiang, Gao and Dong, Guanting and Mou, Yutao and Wang, Liwen and Zeng, Chen and Guo, Daichi and Sun, Mingyang and Xu, Weiran},
  booktitle={Proceedings of the 2022 Conference on Empirical Methods in Natural Language Processing},
  pages={2460--2465},
  year={2022}
}

@article{finch2023leveraging,
  title={Leveraging large language models for automated dialogue analysis},
  author={Finch, Sarah E and Paek, Ellie S and Choi, Jinho D},
  journal={arXiv preprint arXiv:2309.06490},
  year={2023}
}

@inproceedings{sabour2022cem,
  title={Cem: Commonsense-aware empathetic response generation},
  author={Sabour, Sahand and Zheng, Chujie and Huang, Minlie},
  booktitle={Proceedings of the AAAI Conference on Artificial Intelligence},
  volume={36},
  number={10},
  pages={11229--11237},
  year={2022}
}

@article{touvron2023llama,
  title={Llama: Open and efficient foundation language models},
  author={Touvron, Hugo and Lavril, Thibaut and Izacard, Gautier and Martinet, Xavier and Lachaux, Marie-Anne and Lacroix, Timoth{\'e}e and Rozi{\`e}re, Baptiste and Goyal, Naman and Hambro, Eric and Azhar, Faisal and others},
  journal={arXiv preprint arXiv:2302.13971},
  year={2023}
}

@article{reid2024gemini,
  title={Gemini 1.5: Unlocking multimodal understanding across millions of tokens of context},
  author={Reid, Machel and Savinov, Nikolay and Teplyashin, Denis and Lepikhin, Dmitry and Lillicrap, Timothy and Alayrac, Jean-baptiste and Soricut, Radu and Lazaridou, Angeliki and Firat, Orhan and Schrittwieser, Julian and others},
  journal={arXiv preprint arXiv:2403.05530},
  year={2024}
}

@article{liu2023pre,
  title={Pre-train, prompt, and predict: A systematic survey of prompting methods in natural language processing},
  author={Liu, Pengfei and Yuan, Weizhe and Fu, Jinlan and Jiang, Zhengbao and Hayashi, Hiroaki and Neubig, Graham},
  journal={ACM Computing Surveys},
  volume={55},
  number={9},
  pages={1--35},
  year={2023},
  publisher={ACM New York, NY}
}

@article{gao2020making,
  title={Making pre-trained language models better few-shot learners},
  author={Gao, Tianyu and Fisch, Adam and Chen, Danqi},
  journal={arXiv preprint arXiv:2012.15723},
  year={2020}
}

@article{li2021prefix,
  title={Prefix-tuning: Optimizing continuous prompts for generation},
  author={Li, Xiang Lisa and Liang, Percy},
  journal={arXiv preprint arXiv:2101.00190},
  year={2021}
}

@article{cohen1960coefficient,
  title={A coefficient of agreement for nominal scales},
  author={Cohen, Jacob},
  journal={Educational and psychological measurement},
  volume={20},
  number={1},
  pages={37--46},
  year={1960},
  publisher={Sage Publications Sage CA: Thousand Oaks, CA}
}

@article{artstein2008inter,
  title={Inter-coder agreement for computational linguistics},
  author={Artstein, Ron and Poesio, Massimo},
  journal={Computational linguistics},
  volume={34},
  number={4},
  pages={555--596},
  year={2008},
  publisher={MIT Press One Rogers Street, Cambridge, MA 02142-1209, USA journals-info~…}
}

@article{spooren2010coding,
  title={Coding coherence relations: Reliability and validity},
  author={Spooren, Wilbert and Degand, Liesbeth},
  year={2010},
  publisher={Walter de Gruyter GmbH \& Co. KG}
}

@article{achiam2023gpt,
  title={Gpt-4 technical report},
  author={Achiam, Josh and Adler, Steven and Agarwal, Sandhini and Ahmad, Lama and Akkaya, Ilge and Aleman, Florencia Leoni and Almeida, Diogo and Altenschmidt, Janko and Altman, Sam and Anadkat, Shyamal and others},
  journal={arXiv preprint arXiv:2303.08774},
  year={2023}
}

@book{schwartz2016abcs,
  title={The ABCs of how we learn: 26 scientifically proven approaches, how they work, and when to use them},
  author={Schwartz, Daniel L and Tsang, Jessica M and Blair, Kristen P},
  year={2016},
  publisher={WW Norton \& Company}
}

@article{liu2019roberta,
  title={Roberta: A robustly optimized bert pretraining approach},
  author={Liu, Yinhan and Ott, Myle and Goyal, Naman and Du, Jingfei and Joshi, Mandar and Chen, Danqi and Levy, Omer and Lewis, Mike and Zettlemoyer, Luke and Stoyanov, Veselin},
  journal={arXiv preprint arXiv:1907.11692},
  year={2019}
}

@inproceedings{devlin2019bert,
  title={Bert: Pre-training of deep bidirectional transformers for language understanding},
  author={Devlin, Jacob and Chang, Ming-Wei and Lee, Kenton and Toutanova, Kristina},
  booktitle={Proceedings of the 2019 conference of the North American chapter of the association for computational linguistics: human language technologies, volume 1 (long and short papers)},
  pages={4171--4186},
  year={2019}
}

@article{dubey2024llama,
  title={The llama 3 herd of models},
  author={Dubey, Abhimanyu and Jauhri, Abhinav and Pandey, Abhinav and Kadian, Abhishek and Al-Dahle, Ahmad and Letman, Aiesha and Mathur, Akhil and Schelten, Alan and Yang, Amy and Fan, Angela and others},
  journal={arXiv e-prints},
  pages={arXiv--2407},
  year={2024}
}

@article{wei2021finetuned,
  title={Finetuned language models are zero-shot learners},
  author={Wei, Jason and Bosma, Maarten and Zhao, Vincent Y and Guu, Kelvin and Yu, Adams Wei and Lester, Brian and Du, Nan and Dai, Andrew M and Le, Quoc V},
  journal={arXiv preprint arXiv:2109.01652},
  year={2021}
}

@article{chung2024scaling,
  title={Scaling instruction-finetuned language models},
  author={Chung, Hyung Won and Hou, Le and Longpre, Shayne and Zoph, Barret and Tay, Yi and Fedus, William and Li, Yunxuan and Wang, Xuezhi and Dehghani, Mostafa and Brahma, Siddhartha and others},
  journal={Journal of Machine Learning Research},
  volume={25},
  number={70},
  pages={1--53},
  year={2024}
}

@article{tulpan2023deeper,
  title={A Deeper (Autoregressive) Approach to Non-Convergent Discourse Parsing},
  author={Tulpan, Yoav and Tsur, Oren},
  journal={arXiv preprint arXiv:2305.12510},
  year={2023}
}

@inproceedings{li2023task,
  title={Task-aware self-supervised framework for dialogue discourse parsing},
  author={Li, Wei and Zhu, Luyao and Shao, Wei and Yang, Zonglin and Cambria, Erik},
  booktitle={2023 Conference on Empirical Methods in Natural Language Processing (EMNLP 2023)},
  pages={14162--14173},
  year={2023},
  organization={Association for Computational Linguistics}
}

@inproceedings{mim-etal-2022-lpattack,
    title = "{LPA}ttack: A Feasible Annotation Scheme for Capturing Logic Pattern of Attacks in Arguments",
    author = "Mim, Farjana Sultana  and
      Inoue, Naoya  and
      Naito, Shoichi  and
      Singh, Keshav  and
      Inui, Kentaro",
    editor = "Calzolari, Nicoletta  and
      B{\'e}chet, Fr{\'e}d{\'e}ric  and
      Blache, Philippe  and
      Choukri, Khalid  and
      Cieri, Christopher  and
      Declerck, Thierry  and
      Goggi, Sara  and
      Isahara, Hitoshi  and
      Maegaard, Bente  and
      Mariani, Joseph  and
      Mazo, H{\'e}l{\`e}ne  and
      Odijk, Jan  and
      Piperidis, Stelios",
    booktitle = "Proceedings of the Thirteenth Language Resources and Evaluation Conference",
    month = jun,
    year = "2022",
    address = "Marseille, France",
    publisher = "European Language Resources Association",
    url = "https://aclanthology.org/2022.lrec-1.261/",
    pages = "2446--2459"
}

@inproceedings{li2024dialogue,
  title={Dialogue discourse parsing as generation: a sequence-to-sequence LLM-based Approach},
  author={Li, Chuyuan and Yin, Yuwei and Carenini, Giuseppe},
  booktitle={Proceedings of the 25th annual meeting of the special interest group on discourse and dialogue},
  pages={1--14},
  year={2024}
}

@inproceedings{cimino2024coherence,
  title={Coherence-based dialogue discourse structure extraction using open-source large language models},
  author={Cimino, Gaetano and Li, Chuyuan and Carenini, Giuseppe and Deufemia, Vincenzo},
  booktitle={Proceedings of the 25th Annual Meeting of the Special Interest Group on Discourse and Dialogue},
  pages={297--316},
  year={2024}
}

@inproceedings{gu2021dialogbert,
  title={Dialogbert: Discourse-aware response generation via learning to recover and rank utterances},
  author={Gu, Xiaodong and Yoo, Kang Min and Ha, Jung-Woo},
  booktitle={Proceedings of the AAAI Conference on Artificial Intelligence},
  volume={35},
  number={14},
  pages={12911--12919},
  year={2021}
}

@article{mim2021corruption,
  title={Corruption is not all bad: Incorporating discourse structure into pre-training via corruption for essay scoring},
  author={Mim, Farjana Sultana and Inoue, Naoya and Reisert, Paul and Ouchi, Hiroki and Inui, Kentaro},
  journal={IEEE/ACM Transactions on Audio, Speech, and Language Processing},
  volume={29},
  pages={2202--2215},
  year={2021},
  publisher={IEEE}
}

@article{alic2022computationally,
  title={Computationally identifying funneling and focusing questions in classroom discourse},
  author={Alic, Sterling and Demszky, Dorottya and Mancenido, Zid and Liu, Jing and Hill, Heather and Jurafsky, Dan},
  journal={arXiv preprint arXiv:2208.04715},
  year={2022}
}

@article{guo2025deepseek,
  title={Deepseek-r1: Incentivizing reasoning capability in llms via reinforcement learning},
  author={Guo, Daya and Yang, Dejian and Zhang, Haowei and Song, Junxiao and Zhang, Ruoyu and Xu, Runxin and Zhu, Qihao and Ma, Shirong and Wang, Peiyi and Bi, Xiao and others},
  journal={arXiv preprint arXiv:2501.12948},
  year={2025}
}

@inproceedings{jensen2021deep,
  title={A deep transfer learning approach to modeling teacher discourse in the classroom},
  author={Jensen, Emily and L. Pugh, Samuel and K. D'Mello, Sidney},
  booktitle={LAK21: 11th international learning analytics and knowledge conference},
  pages={302--312},
  year={2021}
}

@article{radford2018improving,
  title={Improving language understanding by generative pre-training},
  author={Radford, Alec and Narasimhan, Karthik and Salimans, Tim and Sutskever, Ilya and others},
  year={2018},
  publisher={San Francisco, CA, USA}
}

@article{vaswani2017attention,
  title={Attention is all you need},
  author={Vaswani, Ashish and Shazeer, Noam and Parmar, Niki and Uszkoreit, Jakob and Jones, Llion and Gomez, Aidan N and Kaiser, {\L}ukasz and Polosukhin, Illia},
  journal={Advances in neural information processing systems},
  volume={30},
  year={2017}
}

@article{brown2020language,
  title={Language models are few-shot learners},
  author={Brown, Tom and Mann, Benjamin and Ryder, Nick and Subbiah, Melanie and Kaplan, Jared D and Dhariwal, Prafulla and Neelakantan, Arvind and Shyam, Pranav and Sastry, Girish and Askell, Amanda and others},
  journal={Advances in neural information processing systems},
  volume={33},
  pages={1877--1901},
  year={2020}
}

@article{hu2022lora,
  title={Lora: Low-rank adaptation of large language models.},
  author={Hu, Edward J and Shen, Yelong and Wallis, Phillip and Allen-Zhu, Zeyuan and Li, Yuanzhi and Wang, Shean and Wang, Lu and Chen, Weizhu and others},
  journal={ICLR},
  volume={1},
  number={2},
  pages={3},
  year={2022}
}

@inproceedings{jacob2018quantization,
  title={Quantization and training of neural networks for efficient integer-arithmetic-only inference},
  author={Jacob, Benoit and Kligys, Skirmantas and Chen, Bo and Zhu, Menglong and Tang, Matthew and Howard, Andrew and Adam, Hartwig and Kalenichenko, Dmitry},
  booktitle={Proceedings of the IEEE conference on computer vision and pattern recognition},
  pages={2704--2713},
  year={2018}
}

@article{tang2025few,
  title={The few-shot dilemma: Over-prompting large language models},
  author={Tang, Yongjian and Tuncel, Doruk and Koerner, Christian and Runkler, Thomas},
  journal={arXiv preprint arXiv:2509.13196},
  year={2025}
}

@article{barnett2024fine,
  title={Fine-tuning or fine-failing? debunking performance myths in large language models},
  author={Barnett, Scott and Brannelly, Zac and Kurniawan, Stefanus and Wong, Sheng},
  journal={arXiv preprint arXiv:2406.11201},
  year={2024}
}


\begin{IEEEbiography}
    [{\includegraphics[width=1in,height=1.25in,keepaspectratio]{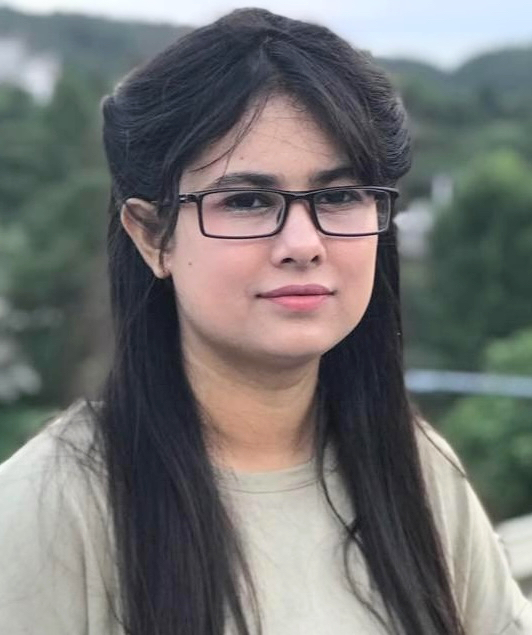}}]{Farjana Sultana Mim} received her B.Sc. in Computer Science and Engineering from Patuakhali Science and Technology University, Bangladesh, in 2016, and her M.S. and Ph.D. in System Information Sciences from Tohoku University, Japan, in 2019 and 2022. She was a postdoctoral scholar in the Department of Electrical and Computer Engineering at Tufts University, USA, in 2023. She is currently a lecturer in the Department of Computer Science and Information Technology at Patuakhali Science and Technology University. Her research interests include NLP in education, large language models, discourse analysis, unsupervised learning, argumentation, and commonsense reasoning.
\end{IEEEbiography}

\vspace*{-.05cm}

\begin{IEEEbiography}
    [{\includegraphics[width=1in,height=1.25in,keepaspectratio]{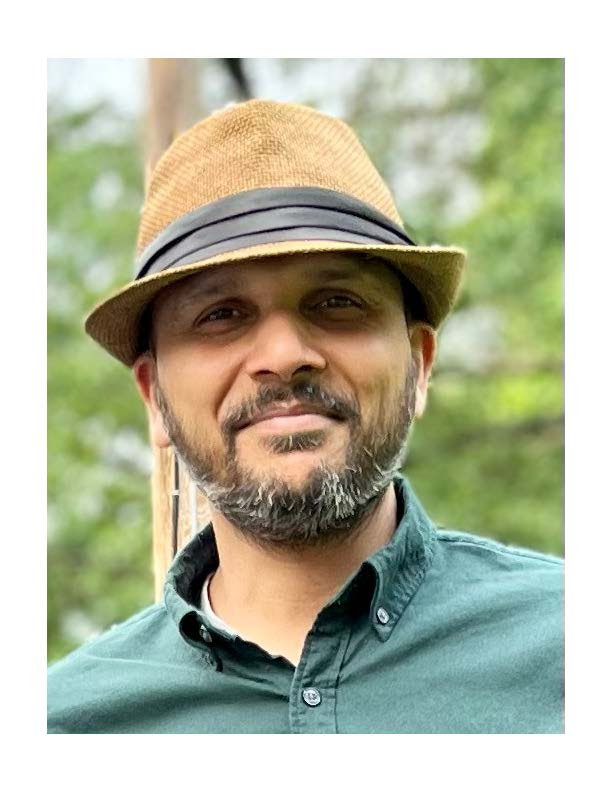}}]{Shuchin Aeron} is a professor in the Department of Electrical and Computer Engineering at Tufts School of Engineering. He received his Ph.D. from Boston University in 2009 and was awarded the best PhD thesis award from both the School Of Engineering and from the Department of Electrical and Computer Engineering. From 2009-2011, he was a postdoctoral research fellow at Schlumberger-Doll Research (SDR), where he worked on signal processing solution products for borehole acoustics resulting in a number of patents. In 2016, he received the NSF CAREER award for his work on multidimensional signals and systems. Shuchin Aeron is presently a senior member of the Institute of Electrical and Electronics Engineers (IEEE) and an associate editor for the ACM transactions on Theory of Probababilistic Machine Learning. 
\end{IEEEbiography}


\begin{IEEEbiography}
    [{\includegraphics[width=1in,height=1.25in,keepaspectratio]{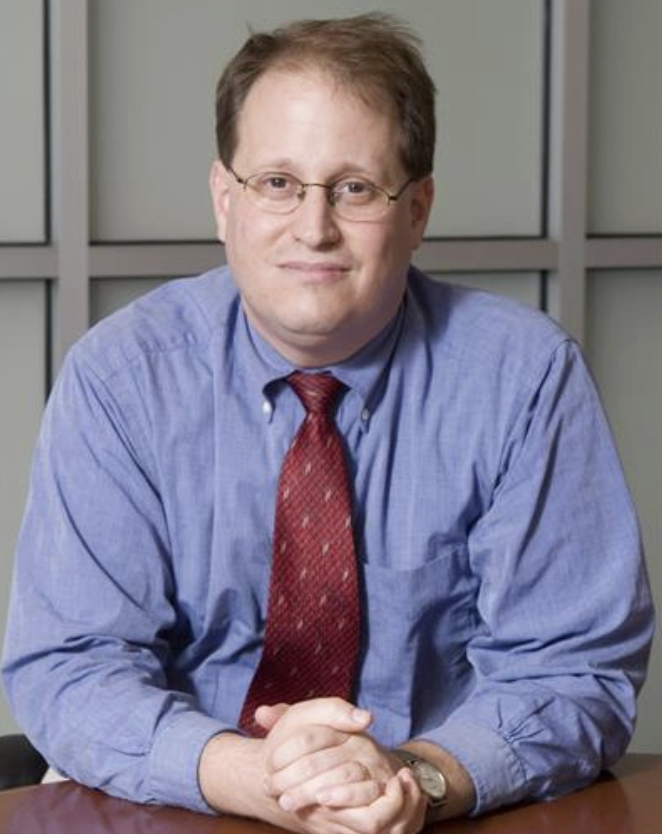}}]{Eric Miller} is a Professor in the Department of Electrical and Computer Engineering and an adjunct Professor in the Departments of Computer Science and Biomedical Engineering at Tufts University. He previously served in the Department of Electrical and Computer Engineering at Northeastern University from 1994 to 2006. He is also a Senior Scientist at the Jean Meyer Human Nutrition Research Center on Aging at Tufts University and currently serves as the Director of the Engineering Education and Centers Division in the Directorate for Engineering at the U.S. National Science Foundation. Dr. Miller received National Science Foundation CAREER Award in 1996 and the Outstanding Research Award from the Northeastern University College of Engineering in 2002. From 2014 to 2018, he served on the Technical Liaison Committee for the IEEE Transactions on Computational Imaging and chaired the SIAM Imaging Sciences Special Interest Group from 2015 to 2017. He was an Associate Editor for the IEEE Transactions on Geoscience and Remote Sensing from 2003 to 2015 and for the IEEE Transactions on Image Processing from 1999 to 2003.
\end{IEEEbiography}

\begin{IEEEbiography}
    [{\includegraphics[width=1in,height=1.25in,keepaspectratio]{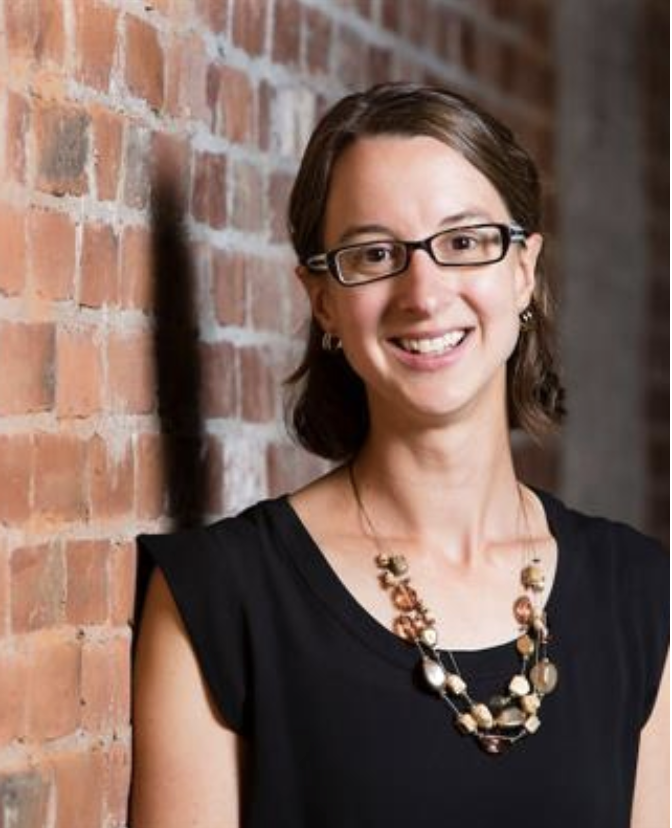}}]{Kristen Wendell} is an Associate Professor in the Department of Mechanical Engineering and Education at Tufts University. She earned her B.S.E. from Princeton University, her M.S. from the Massachusetts Institute of Technology, and her Ph.D. from Tufts University in 2003, 2005, and 2011, respectively. She currently serves as a CEEO Fellow in the Center for Engineering Education Outreach and as the Co-Director of the Institute for Research on Learning and Instruction at Tufts University. Her research work focuses on characterizing and supporting inclusive, sophisticated disciplinary practices during engineering learning experiences in undergraduate course, K-8 classrooms, and teacher education contexts.
\end{IEEEbiography}

 




\vfill

\end{document}